\documentclass[runningheads]{llncs}

\usepackage{graphicx}
\usepackage{amsmath}
\usepackage{amssymb}
\usepackage{booktabs} 
\usepackage{multirow}
\usepackage[colorlinks=true,linkcolor=blue,citecolor=blue,urlcolor=blue]{hyperref}
\usepackage[capitalize]{cleveref}

\begin{document}

\title{Pain in 3D: Controllable Generation of Synthetic Faces for Automated Pain Assessment}
\titlerunning{Pain in 3D}

\author{Xin Lei Lin\inst{1,2,3,4}\thanks{These authors contributed equally to this work.} \and
Soroush Mehraban\inst{1,2,5}\fnmsep\protect\footnotemark[1] \and
Abhishek Moturu\inst{1,2,3} \and
Babak Taati\inst{1,2,3}}
\authorrunning{Lin et al.}

\institute{Vector Institute \and
KITE Research Institute, University Health Network \and
Department of Computer Science, University of Toronto \and
Department of Medical Imaging, University of Toronto \and
Institute of Biomedical Engineering, University of Toronto\\
\email{xinlei.lin@mail.utoronto.ca, soroush.mehraban@mail.utoronto.ca, moturuab@cs.toronto.edu, babak.taati@uhn.ca}}
\maketitle

\begin{abstract}
Automated pain assessment from facial expressions is crucial for non-communicative patient. Progress has been limited by two challenges: (i) existing datasets exhibit severe demographic and label imbalance due to ethical constraints, and (ii) current generative models cannot precisely control facial action units (AUs), facial structure, or clinically validated pain levels.

We introduce \textbf{3DPain}, a large-scale synthetic dataset designed to overcome data scarcity in automated pain assessment. Comprising 82,500 frames across 2,500 unique identities, 3DPain offers extensive heterogeneity in facial pain responses across demographic groups balanced by age, gender, and ethnicity.Our three-stage framework samples diverse 3D meshes, textures them with diffusion models, and applies AU-driven face rigging to synthesize multi-view faces with paired neutral/pain images, facial action units, PSPI scores, and pain-region heatmaps.

We further introduce \textbf{ViTPain}, a Vision Transformer–based framework leveraging cross-attention with a neutral reference face to achieve identity-aware pain estimation. Together, 3DPain and ViTPain establish a controllable, diverse, and clinically grounded foundation for generalizable automated pain assessment.

\keywords{Pain Detection \and Facial Expression Recognition \and Diffusion Models \and Distillation Models \and Generative AI.}
\end{abstract}

\begin{figure}[t]
  \centering
  \includegraphics[width=\textwidth]{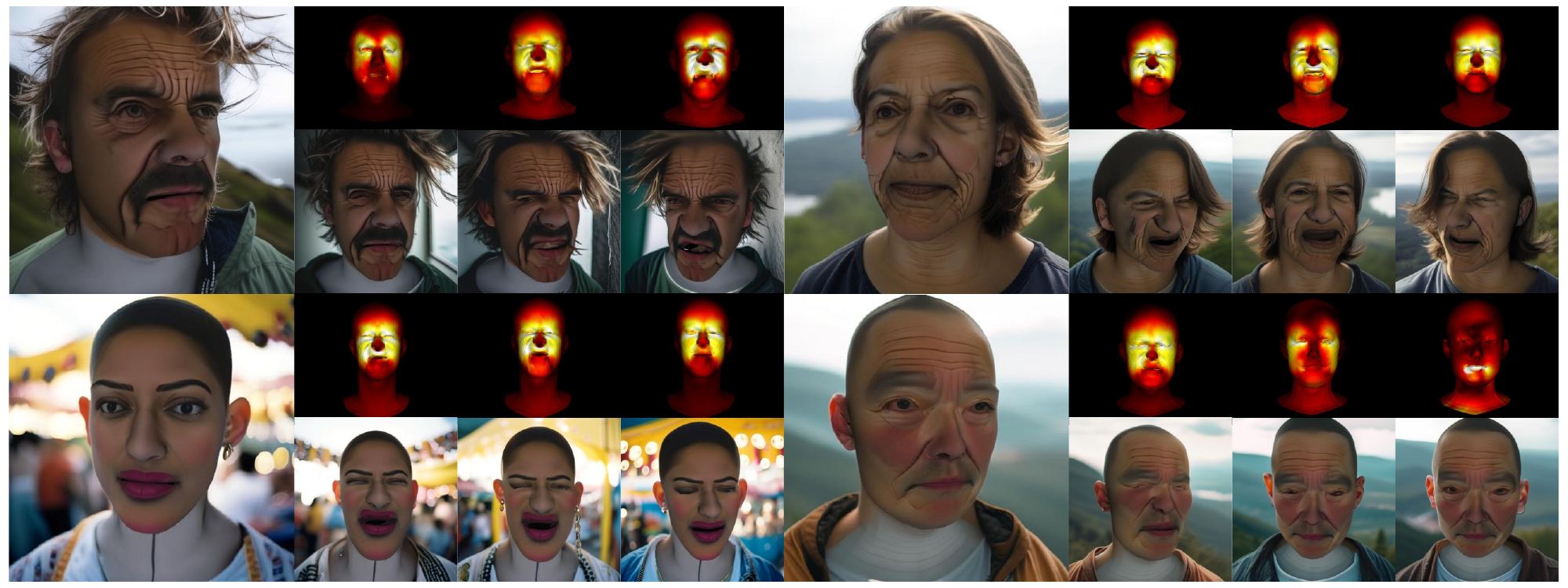} 
  \caption{\textbf{3DPain synthetic dataset.} Our dataset provides multiview images and corresponding heatmaps of facial expressions, enabling robust training and evaluation for automatic pain recognition.}
  \label{fig:teaser}
\end{figure}

\section{Introduction}
\label{sec:intro}

%% NEED TO CONTINUE ON THE FACT THAT THEY ARE PHOTOREALISTIC AND VERY ACCURATE< BUT DIFFICULTY in CONDITIONING into precise outputs.
Automated pain assessment from facial expressions is a critical technology for monitoring vulnerable populations, particularly patients with dementia, severe cognitive impairments, or communication disabilities \cite{Hadjistavropoulos2014,Lucey2011}. While pain is a subjective experience, facial expressions serve as observable behavioral markers that are strongly correlated with pain intensity \cite{Prkachin2008}. However, leveraging these cues for robust computer vision systems is stalled by a fundamental bottleneck: the extreme scarcity of diverse, high-intensity pain data required to train subject-independent models \cite{Rezaei2021}.

 Traditional collection of pain data is invasive, and it is ethically infeasible to induce severe pain in diverse participants to create balanced datasets \cite{Herr2011,Hadjistavropoulos2014}. Consequently, current benchmarks like the UNBC-McMaster dataset \cite{Lucey2011} are small and suffer from severe class imbalance, where frames exhibiting high-intensity pain are rare outliers. Models trained on such limited data frequently overfit to identity or specific demographics, failing to generalize to unseen populations, which is a critical failure mode for clinical deployment \cite{Buolamwini2018,Obermeyer2019}.

Diffusion models have transformed generative computer vision by enabling the photorealistic synthesis of complex data through iterative denoising processes \cite{ho2020denoisingdiffusionprobabilisticmodels}, demonstrating remarkable capabilities across domains ranging from text-to-image synthesis to 3D content creation \cite{rombach2022high,guo2024maisi}. Yet, for clinical pain assessment, visual fidelity alone is insufficient. Standard 2D conditioning methods and prompt-based guidance often produce visually plausible but clinically invalid outputs that do not capture the specific muscle activation patterns defined by the Facial Action Coding System (FACS). For instance, methods like ILVR \cite{choi2021ilvrconditioningmethoddenoising} and SynPain \cite{Taati2025} can generate realistic "pain-like" expressions, but lack the mechanisms to target specific Facial Action Units (AUs) or guarantee alignment with the Prkachin and Solomon Pain Intensity (PSPI) metric.

\paragraph{Threefold Contribution.} Thus, our contributions are threefold:
\begin{enumerate}
\item We introduce a three-stage controllable generation pipeline: (1) FLAME-based neutral face generation via depth-conditioned Kandinsky 2.2 \cite{Li2017FLAME,KandinskyDataloop}, (2) Hunyuan3D 2.1 texture synthesis \cite{hunyuan3d2025hunyuan3d}, and (3) neural face rigging with AU-precise pain expressions and multi-viewpoint rendering \cite{Qin2023NeuralFaceRigging}. This approach enables precise control over facial structures, action units and pain expression levels while maintaining diverse and realistic texture.
\item We release the 3DPain dataset, comprising 82,500 frames across 25,000 pain expressions, and 2,500 synthetic identities generated with diverse ethnic/racial groups, genders, and age ranges, each annotated with exact AU configurations and PSPI scores \cite{Prkachin2008}.
% \item We demonstrate a distillation-based PSPI classification model that leverages facial heatmaps as teaching signals \cite{wei2025magicfacehighfidelityfacialexpression} to improve both accuracy and interpretability of pain predictions.

% TO REPHRASE
\item We demonstrate a reference-based Vision Transformer (ViT) for PSPI regression and pain assessment that leverages cross-attended neutral-pain image pairs to better isolate dynamic pain expressions from static identity features.
\end{enumerate}

\section{Related Works}
\label{sec:related_works}
\paragraph{Facial Expression Framework for Pain Assessment.} Facial expressions serve as a critical window into pain experience, offering physiological indicators that transcend language barriers \cite{Ekman1978,Hammal2012}. The Facial Action Coding System (FACS) quantifies facial movements using Action Units (AUs), where higher AU values indicate stronger or more pronounced facial movements, which are closely correlated with pain assessments \cite{Ekman1978}. The Prkachin and Solomon Pain Index (PSPI) represents the gold standard for objective facial pain evaluation. It aggregates AUs associated with core pain responses: brow lowering ($\text{AU}_4$), orbital tightening ($\text{AU}_6$ and $\text{AU}_7$), levator contraction around the nose and upper lip ($\text{AU}_9$ and $\text{AU}_{10}$), and eye closure ($\text{AU}_{43}$). The index is defined as:
% \begin{align}
% \text{PSPI}
%     = \underbrace{\text{AU}_4}_{\text{Brow Lowerer}}
%     + \max(\underbrace{\text{AU}_6}_{\text{Cheek Raiser}}, \underbrace{\text{AU}_7}_{\text{Lid Tightener}}) \nonumber \\
%     + \max(\underbrace{\text{AU}_9}_{\text{Nose Wrinkler}}, \underbrace{\text{AU}_{10}}_{\text{Upper Lip Raiser}})
%     + \underbrace{\text{AU}_{43}}_{\text{Eyes Closed}}
% \label{eq:pspi}
% \end{align}
\begin{equation}
\resizebox{0.93\hsize}{!}{$
\text{PSPI} = \underbrace{\text{AU}_4}_{\text{Brow Lowerer}} + \max(\underbrace{\text{AU}_6}_{\text{Cheek Raiser}}, \underbrace{\text{AU}_7}_{\text{Lid Tightener}}) + \max(\underbrace{\text{AU}_9}_{\text{Nose Wrinkler}}, \underbrace{\text{AU}_{10}}_{\text{Upper Lip Raiser}}) + \underbrace{\text{AU}_{43}}_{\text{Eyes Closed}}
$}
\label{eq:pspi}
\end{equation}
% IF NOT ENOUGH SPACE USE THIS:
% \begin{align}
% \text{PSPI}
%     &= \text{AU}_4
%     + \max(\text{AU}_6, \text{AU}_7) \nonumber + \max(\text{AU}_9, \text{AU}_{10})
%     + \text{AU}_{43}
% \label{eq:pspi}
% \end{align}
This formulation is strongly correlated with self-reported pain scores across diverse populations where higher PSPI corresponds to more intense pain \cite{Prkachin2008}.

\paragraph{Dataset Limitations and Bias.}
Despite the clinical relevance of PSPI, existing facial pain datasets are severely limited. The UNBC-McMaster Shoulder Pain dataset remains the most widely used benchmark, yet it contains only 25 participants with narrow demographic diversity, leading to underrepresentation across ethnicity, race, and age groups \cite{Lucey2011}. Moreover, high-intensity pain expressions (high PSPI scores) are extremely rare, since it is ethically not feasible to induce or record severe pain in controlled environments. This scarcity creates strong class imbalance and limits model robustness for recognizing clinically critical states. Other datasets like BioVid attempt stimulus-based labeling \cite{walter2013biovid}, but stimulus intensity does not necessarily align with facial expressions or subjective pain \cite{Prkachin2008}. FACS datasets outside of pain research often lack pain-related facial expressions entirely \cite{Zhang_2016_CVPR}, while many clinical datasets remain private due to ethical and privacy concerns \cite{Rezaei2021}. These limitations collectively hinder the development of scalable, unbiased, and generalizable pain recognition models.
\paragraph{Synthetic Data for Vision Tasks.}
Synthetic data has emerged as a promising solution for addressing dataset scarcity, imbalance, and demographic bias in vision tasks. Prior work shows its effectiveness in face-analysis \cite{fake2021}, 3D human mesh recovery \cite{bedlam2023}, and identity-preserving face generation \cite{idiff2023}. These results suggest that well-designed synthetic data pipelines can significantly improve model robustness and generalization, especially in domains where large-scale annotated real data is infeasible to collect.

\paragraph{Facial Generation with Action Unit Control.} Recent advancements have focused on enabling fine-grained control over facial Action Units (AUs) to synthesize precise expressions, yet significant barriers remain for clinical application. While approaches such as Liu et al. \cite{LIU2019200} leverage 3D canonicalization for AU-conditioned synthesis, they rely on 2D training paradigms that restrict multi-view robustness. Critically, data scarcity and supervision limitations in these frameworks create two distinct bottlenecks for pain modeling. First, standard training datasets (e.g., DISFA, BP4D) lack annotations for the Lid Tightener (AU$_7$) and Eyes Closed (AU$_{43}$), leading to the direct omission of these essential components of the Prkachin and Solomon Pain Intensity (PSPI) metric \cite{LIU2019200}. Second, reliance on geometric approximations struggles to capture high-frequency details; specifically, the Cheek Raiser (AU$_6$) and Nose Wrinkler (AU$_9$) are often rendered with low fidelity because they depend on subtle, transient shape deformations that are difficult to distinctively render without explicit 3D surface guidance \cite{LIU2019200}. Other audio-driven methods \cite{Fan2022FaceFormer} face similar limitations, prioritizing lip-sync over anatomical fidelity. Furthermore, methods utilizing GAN inversion for attribute manipulation \cite{Yin2022StyleHeat,c2a2_2024} face an ill-posed inverse problem, creating an inherent trade-off between reconstruction fidelity and identity preservation \cite{xia2022gan,tov2021designing}. Similarly, while diffusion-based approaches have been applied for identity-preserving synthesis \cite{wei2025magicfacehighfidelityfacialexpression}, they rely on prompt-driven guidance that lacks the granularity to explicitly target Action Units or enforce clinical metrics \cite{Taati2025,sohldickstein2015deepunsupervisedlearningusing}. Since these models cannot inherently guarantee the validity of the generated expressions, they depend entirely on costly post-hoc manual annotation, creating a significant bottleneck that restricts the availability of large-scale, clinically validated training data \cite{Taati2025}.

\begin{figure}[t]
  \centering
  \includegraphics[width=\textwidth]{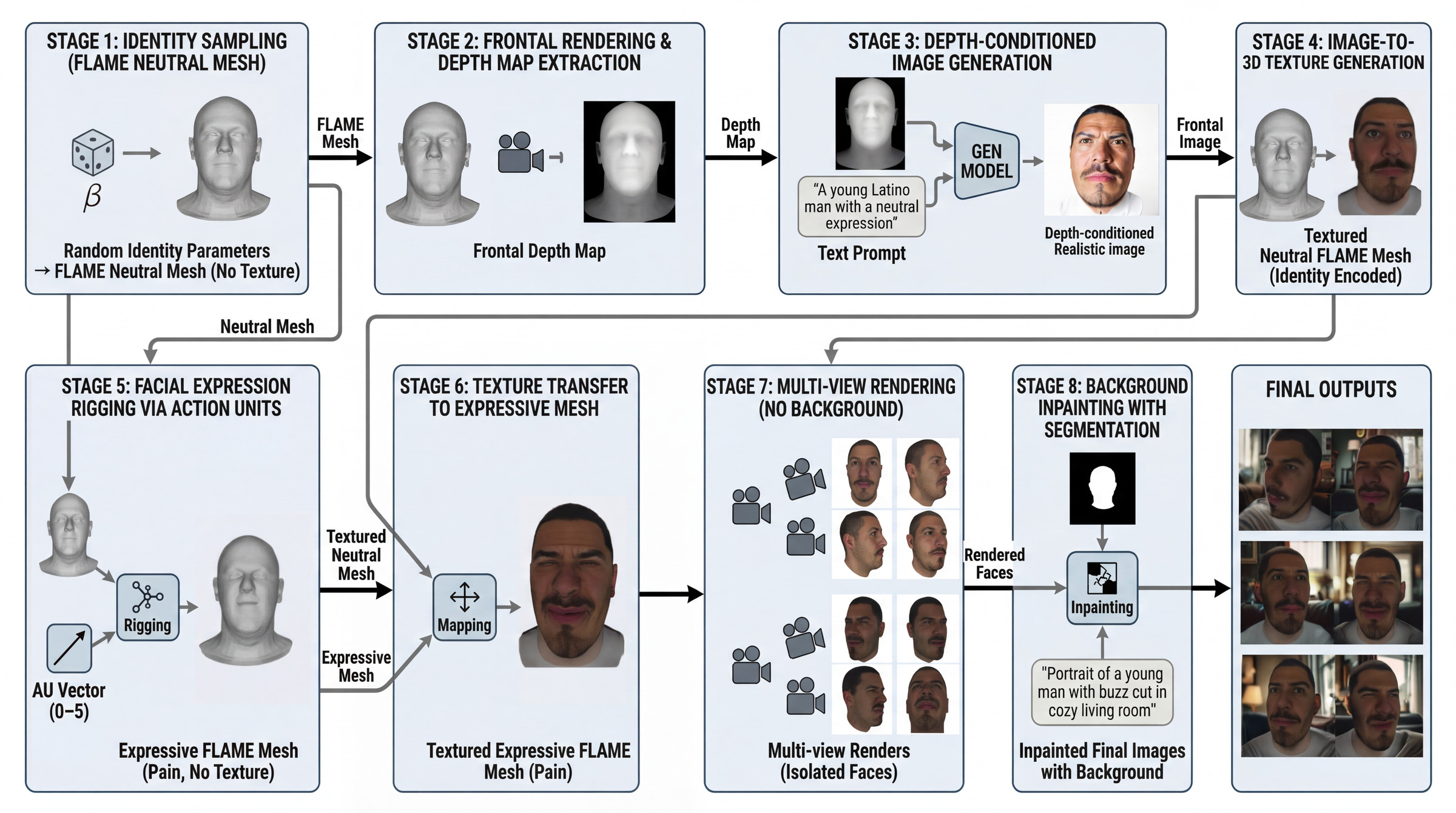}
  \caption{\textbf{Pipeline for facial pain expression generation.} A neutral FLAME identity is sampled and rendered as a frontal depth map, which conditions a depth-guided image generation model to synthesize a realistic face. The generated image is used to texture the neutral mesh, after which pain expressions are introduced via Action Unit–based rigging and texture transfer. The expressive mesh is rendered from multiple views and refined with segmentation-based background inpainting to produce the final images.}
  \label{fig:faces_pain}
\end{figure}

\section{Data Generation}
\label{sec:data_generation}

To address the limited availability of high-quality pain expression datasets, we develop a comprehensive generative pipeline that produces synthetic facial expressions with precise control over pain-related action units (AUs). Our approach combines state-of-the-art 3D facial modeling with diffusion-based image synthesis to create a diverse, PSPI-grounded dataset for automated pain assessment training. The resulting dataset, 3DPain, comprises 82,500 frames derived from 2,500 synthetic identities across a wide spectrum of ages, genders, and ethnicities. Crucially, each subject is rendered from multiple viewpoints, capturing a full range of neutral and pain-induced expressions.

\paragraph{Diffusion-Based Synthesis Framework.} We leverage latent diffusion models (LDMs) \cite{rombach2022high} to bridge the gap between coarse 3D anatomical constraints and photorealistic appearance. While facial meshes provide the necessary geometric control for Action Units, they lack the high-frequency surface details (such as skin texture, wrinkles, and hair) required for domain realism. LDMs address this by enabling conditional synthesis that respects the underlying 3D geometry while hallucinating realistic textures. Furthermore, their probabilistic formulation ensures extensive mode coverage, which is critical for generating a wide range of demographic attributes (age, ethnicity, gender).

% No need diffusion fluff
% During training, the forward diffusion process transforms data $x_0$ from the original distribution into pure noise over $T$ timesteps according to:
% \begin{align}
% q(x_t \mid x_{t-1}) = \mathcal{N}\big(x_t; \sqrt{1-\beta_t} x_{t-1}, \beta_t I \big),
% \end{align}
% where $\beta_t$ is a variance schedule controlling the noise injection rate. The model learns to reverse this process by predicting the noise $\epsilon$ added at each timestep through a neural network $\epsilon_\theta(x_t, t)$, optimizing the following objective:
% \begin{align}
% L = \mathbb{E}_{x_0, \epsilon, t} \big[\|\epsilon - \epsilon_\theta(x_t, t)\|_2^2 \big],
% \end{align}
% where $x_t$ represents the noisy version of the original data $x_0$ at timestep $t$.

\paragraph{Multi-Modal Conditional Generation.}
In our pipeline, conditional diffusion models serve three critical roles: (1), depth-conditioned generation produces neutral facial images from FLAME-derived depth maps \cite{Li2017FLAME}, ensuring geometric consistency between 3D mesh structure and 2D photorealistic appearance; (2), 3D texture synthesis maps realistic physically based rendering textures onto FLAME meshes \cite{hunyuan3d2025hunyuan3d} to capture fine-grained skin details and ethnic features; (3), inpainting-based diffusion handles background synthesis and hair completion \cite{KandinskyDataloop} to produce natural-looking synthetic faces (see \cref{fig:faces_pain}).

Our multi-stage generative pipeline therefore addresses three key challenges in synthetic pain data generation: (i) maintaining anatomical consistency across diverse facial structures, (ii) achieving precise control over relevant pain action units, and (iii) ensuring photorealistic quality suitable for training robust pain detection models. The pipeline integrates complementary generative models, each optimized for specific aspects of facial synthesis while maintaining end-to-end consistency.

\paragraph{3D Meshes for Diverse Facial Identities.}
The FLAME model \cite{Li2017FLAME} provides a parametric 3D mesh framework for representing facial geometry. For instance, FLAME enables the generation of realistic facial structures with varying shapes, allowing us to model individual identity differences while maintaining anatomical plausibility. Furthermore, its 3D representation preserves geometric consistency across viewpoints and lighting conditions, a critical property for reliable pain detection where subtle facial variations must be captured accurately \cite{Ekman1978,Prkachin2008}.

\paragraph{2D Neutral Facial Generation.} We employ Kandinsky 2.2 augmented with ControlNet \cite{KandinskyDataloop} to translate FLAME-derived depth maps into photorealistic neutral faces. By conditioning the latent diffusion process \cite{ho2020denoisingdiffusionprobabilisticmodels,rombach2022high} on these depth maps, we ensure that the synthesized 2D images faithfully adhere to the underlying 3D facial geometry defined by the FLAME shape parameters. Beyond geometric consistency, we leverage the text-promptable nature of the model to enforce demographic diversity. We curated 2,500 unique text prompts spanning all major ethnicities, genders, and age groups (ranging from young adults to the elderly), along with variations in attire. This combination of geometric conditioning and prompt engineering produces a training set that reflects real-world population heterogeneity, thereby enhancing generalizability and mitigating potential biases in downstream pain detection (see \cref{tab:demography}) \cite{Buolamwini2018,Obermeyer2019}.

\paragraph{3D Texture Generation.}
We use the texture generation component of Hunyuan3D 2.1~\cite{hunyuan3d22025tencent} to transform neutral FLAME meshes into realistic physically based rendering (PBR) textures, conditioned on the generated neutral face images. Instead of relying on the default FLAME textures, we leverage Hunyuan3D’s outputs to better capture ethnic features and fine-grained skin details such as wrinkles, leading to a more faithful and diverse representation of facial appearance. This approach builds on the geometric foundation established in the previous stage while producing view-consistent textures that preserve realism under varying camera angles and lighting conditions~\cite{yang2024hunyuan3d}. By mapping geometric features to appropriate PBR texture patterns, the model ensures that the synthesized faces exhibit realistic skin properties, age-appropriate characteristics, and ethnic consistency with the underlying mesh parameters. This stage establishes a high-quality baseline facial appearance before applying pain-specific expressions through neural face rigging.

\paragraph{Neural Face Rigging for Controllable Expressions.}
Neural Face Rigging (NFR) \cite{Qin2023NeuralFaceRigging} enables precise manipulation of facial expressions by learning the mapping between action unit activations and corresponding mesh deformations. Unlike traditional rigging approaches that rely on manually defined control points, NFR uses DiffusionNet~\cite{sharp2022diffusionnet} to extract 3D shape features and Neural Jacobian Fields~\cite{aigerman2022neural} to predict vertex displacements based on desired AU configurations. This approach allows us to generate pain expressions with unprecedented precision by directly targeting specific AUs identified in the PSPI formula (AU$_4$, AU$_6$, AU$_7$, AU$_9$, AU$_{10}$, AU$_{43}$)\cite{Prkachin2008}. The neural rigging process maintains facial identity while applying expression-specific deformations, ensuring that the resulting pain expressions remain anatomically plausible and consistent with clinical observations \cite{Ekman1978,Hadjistavropoulos2014}. By controlling the intensity of individual AUs, we can generate expressions corresponding to specific PSPI scores, enabling the creation of training data with known pain levels.

% \paragraph{Heatmap Generation.} We generate AU heatmaps by comparing the meshes before and after neural face rigging. Specifically, we analyze changes in vertex positions between neutral FLAME meshes and their corresponding pain FLAME meshes to identify which action units have been activated and quantify their intensity. Both the distances between corresponding vertices and the displacements of vertices are used to compute these heatmaps, providing a precise spatial representation of AU activations for downstream training and evaluation.

\paragraph{Kandinsky for Final Background Inpainting and Hair Generation.}
The final stage employs Kandinsky 2.2 \cite{KandinskyDataloop} for background synthesis and hair generation to create complete, photorealistic images. After rendering the textured, rigged 3D faces from multiple viewpoints, we use inpainting techniques \cite{rombach2022high} to generate appropriate backgrounds and complete hair textures that may not be fully captured in the 3D mesh representation. This stage ensures that the synthetic images appear natural and realistic while maintaining focus on the facial region critical for pain detection \cite{Hammal2012,Kunz2017}.
\begin{table}[h]
  \centering
  \caption{Demographic distribution (Total Identities: 2,500).}
  \label{tab:demography}
  \footnotesize
  % 8 columns: Age(2), Gender(2), Ethnicity A(2), Ethnicity B(2)
  \begin{tabular*}{\textwidth}{@{\extracolsep{\fill}} lrlrlrlr}
    \toprule
    \textbf{Age} & \textbf{\#} & \textbf{Gender} & \textbf{\#} & \textbf{Ethnicity} & \textbf{\#} & \textbf{Ethnicity} & \textbf{\#} \\
    \midrule
    Young   & 1,563 & Man   & 1,723 & Latino      & 646 & White      & 460 \\
    Elderly & 937   & Woman & 777   & Middle East & 585 & East Asian & 258 \\
            &       &       &       & South Asian & 469 & Black      & 82  \\
    \bottomrule
  \end{tabular*}
\end{table} % Uncomment if file exists

\begin{figure}[t]
  \centering
  \includegraphics[width=\textwidth]{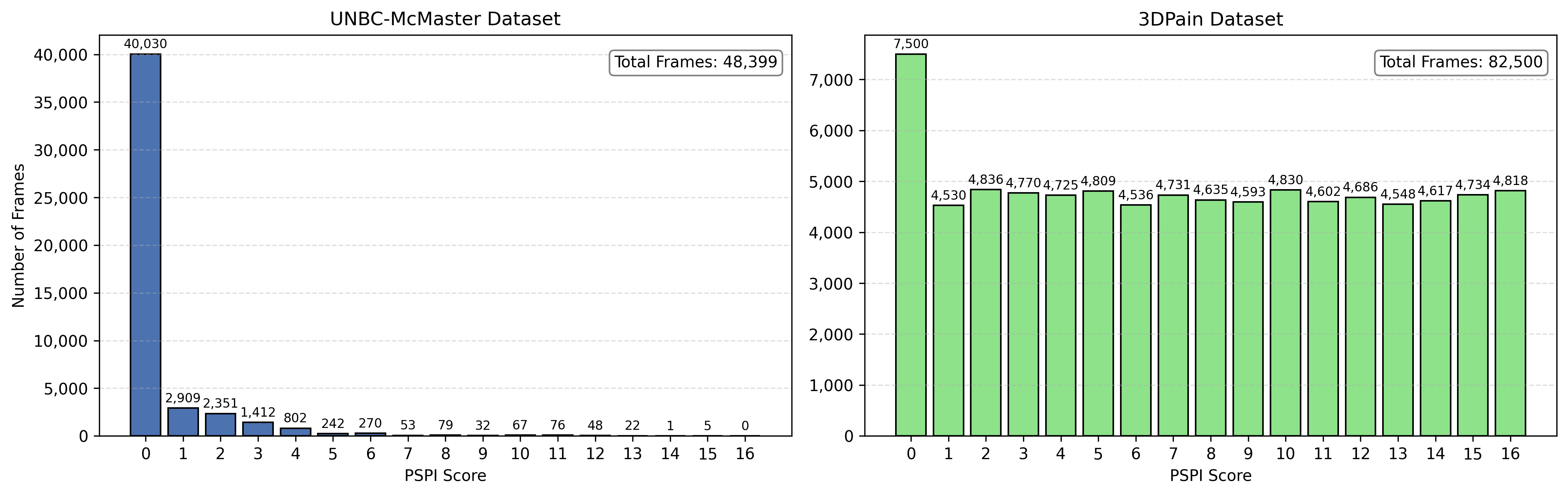}
  \caption{Quantitative comparison of frames in the synthetic 3DPain dataset as well as the UNBC-McMaster dataset for PSPI scores between 0-16.}
  \label{fig:data_distribution}
\end{figure}

\section{Pain Assessment Model Architecture}
\label{sec:pain_assessment_model}
We propose a Vision Transformer architecture \cite{dosovitskiy2020image} for automated pain estimation that addresses the challenges of inter-subject variability and data imbalance. Our approach utilizes the Dino-V3 backbone \cite{simeoni2025dinov3} to extract rich facial representations. To disentangle dynamic pain expressions from static identity features, we integrate a neutral face reference-guided learning paradigm. While previous works have demonstrated the efficacy of pairwise reference learning in Convolutional Neural Networks \cite{Rezaei2021}, we extend this concept to the Vision Transformer landscape via a cross-attention mechanism \cite{vaswani2017attention}. 

\begin{figure}[t]
  \centering
  \includegraphics[width=\textwidth]{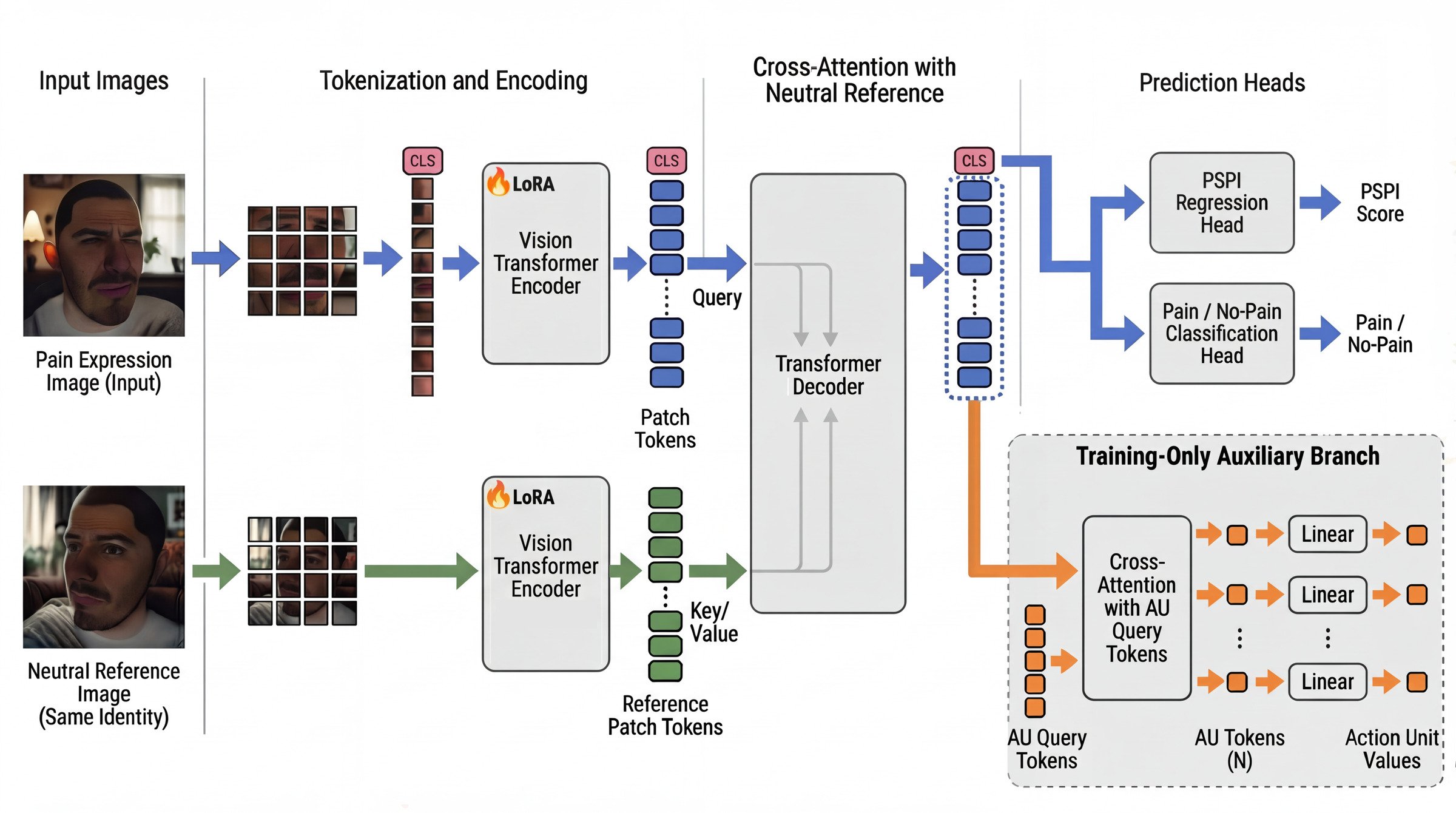}
  \caption{Overview of the proposed ViTPain framework. A pain expression image is tokenized and encoded by a Vision Transformer to produce CLS and patch tokens, which cross-attend to patch tokens from a neutral reference image of the same identity. The resulting CLS token is used for PSPI regression and pain/no-pain classification. During training only, patch tokens additionally cross-attend to learnable AU query tokens to predict Action Unit intensities, providing auxiliary supervision.}
  \label{fig:vitpain}
\end{figure}

\subsection{Neutral Reference Cross-Attention}
The model processes two input images: the target pain image $I_{pain}$ and a reference neutral image $I_{neutral}$ of the same subject. Both are encoded through separate Dino-V3 encoders with distinct LoRA adapters, extracting feature token sequences $\mathbf{Z}_{pain} \in \mathbb{R}^{(N+1) \times D}$ and $\mathbf{Z}_{neutral} \in \mathbb{R}^{(N) \times D}$, where $N$ denotes the number of patches, with one CLS token, and $D$ is the embedding dimension. The pain features then query the neutral reference through an 8-head cross-attention mechanism \cite{vaswani2017attention} where the attention output, $\mathbf{A}_{ref}$, is computed as:
\begin{align}
\mathbf{A}_{ref} &= \text{Softmax}\left(\frac{\mathbf{Q}\mathbf{K}^T}{\sqrt{D}}\right)\mathbf{V}, \\
\text{where \; \;}
\mathbf{Q} = \mathbf{Z}_{pain} \mathbf{W}_Q, \quad
&\mathbf{K} = \mathbf{Z}_{neutral} \mathbf{W}_K, \quad
\mathbf{V} = \mathbf{Z}_{neutral} \mathbf{W}_V \nonumber
\end{align}

We then fuse the aligned neutral reference into the pain representation using a residual connection and layer normalization:
\begin{align}
\mathbf{F}_{context} = \text{LayerNorm}(\mathbf{Z}_{pain} + \mathbf{A}_{ref})
\end{align}

This mechanism enriches the pain representation with identity-specific neutral cues, enabling the subsequent layers to better distinguish dynamic pain deformations from static facial structures. The CLS token from $\mathbf{F}_{context}$ is used for pain score prediction, while the patch tokens are used for Action Unit regression. Finally, we introduce a multi-shot inference strategy in which multiple neutral reference frames are used at test time, and predictions conditioned on each reference are averaged to improve robustness to reference-frame variability.

% \subsection{Query-Based AU Cross-Attention}
% To capture fine-grained muscle movements required for Action Unit regression, a query-based attention module is utilized instead of the global classification token. We introduce learnable AU-specific query tokens $\mathbf{Q}_{AU} \in \mathbb{R}^{6 \times D}$ corresponding to the six pain-relevant AUs ($AU_4$, $AU_6$, $AU_7$, $AU_9$, $AU_{10}$, $AU_{43}$). These queries cross-attend the context patch features $\mathbf{F}_{context}^{patches}$ (obtained after neutral reference cross-attention) to extract localized activation information:

% \begin{align}
% \mathbf{S}_{i,j} &= \frac{\mathbf{Q}_{AU_i} (\mathbf{F}_{context}^{patches})_j^T}{\sqrt{D}} \\
% \alpha_{i,j} &= \frac{\exp(\mathbf{S}_{i,j})}{\sum_{k=1}^{N} \exp(\mathbf{S}_{i,k})} \\
% \mathbf{F}_{AU_i} &= \sum_{j=1}^{N} \alpha_{i,j} (\mathbf{F}_{context}^{patches})_j
% \end{align}

% where $\mathbf{F}_{AU_i}$ represents the attended feature vector specific to AU $i$. This ensures the model extracts spatially distinct features for specific muscle movements, such as brow lowering (AU4) or cheek raising (AU6), which have already been contextualized relative to the subject's neutral baseline.

\subsection{Query-Based AU Cross-Attention}
To capture fine-grained muscle movements, we introduce learnable query tokens $\mathbf{Q}_{AU} \in \mathbb{R}^{6 \times D}$ corresponding to the six pain-relevant AUs. These queries extract localized information from the context patches $\mathbf{F}_{context}^{patches}$ via cross-attention:
\begin{equation}
\mathbf{F}_{AU} = \text{Softmax}\left(\frac{\mathbf{Q}_{AU} (\mathbf{F}_{context}^{patches})^T}{\sqrt{D}}\right)\mathbf{F}_{context}^{patches}
\end{equation}
This yields spatially distinct feature vectors $\mathbf{F}_{AU}$ for specific muscle movements (e.g., brow lowering or cheek raising) contextualized by the neutral baseline.

\subsection{Prediction Heads and Supervision}
The architecture employs a multi-task prediction strategy to robustly estimate pain intensity. The primary supervision is provided by a regression head that maps the CLS token from $\mathbf{F}_{context}$ to a continuous, normalized PSPI score:

\begin{align}
\text{PSPI}_{pred} = \sigma(\text{MLP}_{regression}(\mathbf{F}_{context}^{CLS}))
\end{align}

where $\sigma$ denotes the sigmoid activation to ensure the output remains in the valid range. 

However, regression models often suffer from mean-value collapse due to the prevalence of zero-pain frames in naturalistic datasets. To mitigate this, we incorporate an auxiliary binary classification head trained to distinguish between pain and no-pain states. These auxiliary objectives sharpen the decision boundaries at critical lower PSPIs required for pain classification, preventing the model from predicting negligible intensities for subtle but genuine pain expressions.

For Action Unit prediction, we apply the aforementioned query-based cross-attention mechanism to the patch tokens of $\mathbf{F}_{context}$, producing six independent AU intensity estimates, where ReLU \cite{nair2010rectified} ensures non-negative predictions consistent with AU intensity semantics:

\begin{align}
\text{AU}_i = \text{ReLU}(\text{MLP}_{AU}(\mathbf{F}_{AU_i})), \quad i \in \{4, 6, 7, 9, 10, 43\}
\end{align}

\paragraph{Loss Formulation.} The network is optimized using a composite loss function comprising the regression MSE, the binary classification Cross-Entropy, and the AU regression MSE. We tune the hyperparameters $\lambda$ to roughly ensure that each term contributes equally in magnitude to the final backpropagation gradient:
\begin{align}
\mathcal{L}_{total} &= \lambda_{reg} \mathcal{L}_{MSE}(\hat{y}_{PSPI}, y_{PSPI}) \nonumber \\
& + \lambda_{bin} \mathcal{L}_{BCE}(\hat{y}_{bin}, y_{bin}) \nonumber
+ \lambda_{AU} \mathcal{L}_{MSE}(\hat{\mathbf{y}}_{AU}, \mathbf{y}_{AU})
\end{align}
where $\hat{y}_{PSPI}$ is the predicted PSPI score, $\hat{y}_{bin}$ are the binary classification logits at multiple thresholds, and $\hat{\mathbf{y}}_{AU}$ are the predicted AU intensities.

\section{Experiments}
\label{sec:experiments}
\begin{table}[t]
  \centering
  \caption{\textbf{3DPain Pretraining:} 5-fold CV results on UNBC-McMaster for baselines vs. 3DPain pretraining. Metrics: F1/AUROC (mean $\pm$ std) and aggregated PCC.}
  \label{tab:unbc_results}
  \resizebox{\linewidth}{!}{%
  \setlength{\tabcolsep}{3pt} 
  \begin{tabular}{@{}llccccccc@{}}
    \toprule
    & & \multicolumn{3}{c}{\textbf{F1-Score $\uparrow$}} & \multicolumn{3}{c}{\textbf{AUROC $\uparrow$}} & \multirow{2}{*}{\textbf{PCC $\uparrow$}} \\
    \cmidrule(lr){3-5} \cmidrule(lr){6-8}
    \textbf{Model} & \textbf{Training} & $\ge 1$ & $\ge 2$ & $\ge 3$ & $\ge 1$ & $\ge 2$ & $\ge 3$ & \\
    \midrule
    % MODEL 1: PwCT
    \multirow{2}{*}{\textbf{PwCT \cite{Rezaei2021}}} 
      & Baseline & \underline{0.52} $\pm$ 0.10 & \underline{0.48} $\pm$ 0.16 & 0.45 $\pm$ 0.23 & 0.77 $\pm$ 0.07 & 0.80 $\pm$ 0.09 & 0.80 $\pm$ 0.16 & 0.54 \\
      & W/ 3DPain & \underline{0.52} $\pm$ 0.09 & \underline{0.48} $\pm$ 0.15 & \textbf{0.47} $\pm$ 0.24 & \textbf{0.79} $\pm$ 0.11 & \textbf{0.82} $\pm$ 0.10 & \textbf{0.84} $\pm$ 0.14 & \textbf{0.55} \\
    \midrule
    % MODEL 2: Vanilla ViT
    \multirow{2}{*}{\textbf{Vanilla ViT}} 
      & Baseline & 0.32 $\pm$ 0.08 & 0.24 $\pm$ 0.05 & 0.20 $\pm$ 0.09 & 0.54 $\pm$ 0.14 & 0.57 $\pm$ 0.17 & 0.56 $\pm$ 0.15 & 0.28 \\
      & W/ 3DPain & \textbf{0.46} $\pm$ 0.06 & \textbf{0.41} $\pm$ 0.10 & \textbf{0.40} $\pm$ 0.17 & \textbf{0.73} $\pm$ 0.07 & \textbf{0.76} $\pm$ 0.05 & \textbf{0.85} $\pm$ 0.04 & \textbf{0.44} \\
    \midrule
    % MODEL 3: ViTPain (Ours)
    \multirow{2}{*}{\textbf{ViTPain (Ours)}} 
      & Baseline & 0.47 $\pm$ 0.13 & 0.43 $\pm$ 0.13 & 0.35 $\pm$ 0.17 & 0.75 $\pm$ 0.12 & 0.79 $\pm$ 0.10 & 0.81 $\pm$ 0.06 & 0.36 \\
      & W/ 3DPain & \textbf{0.58} $\pm$ 0.06 & \textbf{0.50} $\pm$ 0.04 & \textbf{0.43} $\pm$ 0.17 & \textbf{0.86} $\pm$ 0.03 & \textbf{0.86} $\pm$ 0.04 & \textbf{0.88} $\pm$ 0.03 & \textbf{0.53} \\
    \bottomrule
  \end{tabular}%
  }
\end{table}

\paragraph{Datasets and Class Imbalance.} 
We utilize 3DPain, a large-scale synthetic dataset with diverse facial pain expressions and precise Action Unit (AU) control, to validate the effectiveness of our pre-training strategy. The UNBC-McMaster Shoulder Pain Expression Archive \cite{Lucey2011} serves as our real-world testbed, containing 48,398 frames from 25 subjects annotated with PSPI scores (0–16). The pain/no-pain ratio in UNBC-McMaster is severely imbalanced, with neutral examples vastly outnumbering high-intensity pain frames. To prevent the model from biasing toward the majority class, we implement a weighted sampling strategy following the protocol established by \cite{Rezaei2021}. Specifically, we assign a fixed weight of 2.0 to samples with PSPI $\geq 1$ (pain) and 1.0 to samples with PSPI $< 1$ (no pain). This ensures that rare high-pain frames are sampled with higher probability, stabilizing the optimization of the ordinal regression heads.

\paragraph{Parameter-Efficient Fine-Tuning.} 
\begin{table*}[t]
  \centering
  \caption{\textbf{Ablation Study:} Step-wise contribution analysis on UNBC-McMaster (5-fold CV) across varying PSPI thresholds. We report metrics (mean $\pm$ std) for single-frame ($k=1$) and multi-shot ($k=3$) configurations.}
  \label{tab:ablation_study}
  
  \resizebox{\textwidth}{!}{%
  \setlength{\tabcolsep}{3pt} 
  \renewcommand{\arraystretch}{0.95} 
  \begin{tabular}{@{}llccccccc@{}}
    \toprule
    & & \multicolumn{3}{c}{\textbf{F1-Score $\uparrow$}} & \multicolumn{3}{c}{\textbf{AUROC $\uparrow$}} & \multirow{2}{*}{\textbf{PCC $\uparrow$}} \\
    \cmidrule(lr){3-5} \cmidrule(lr){6-8}
    \textbf{Model Configuration} & \textbf{$k$-Shot} & $\ge 1$ & $\ge 2$ & $\ge 3$ & $\ge 1$ & $\ge 2$ & $\ge 3$ & \\
    \midrule
    \textbf{1. Vanilla ViT w/ 3D Pain} & - & 0.46 $\pm$ 0.06 & 0.41 $\pm$ 0.10 & 0.40 $\pm$ 0.17 & 0.73 $\pm$ 0.07 & 0.76 $\pm$ 0.05& 0.85 $\pm$ 0.04 & 0.44 \\
    \textbf{2. + AU Query Head} & - & 0.48 $\pm$ 0.04 & 0.44 $\pm$ 0.09 & 0.42 $\pm$ 0.14 & 0.77 $\pm$ 0.08 & 0.79 $\pm$ 0.06 & 0.86 $\pm$ 0.06 & 0.49 \\
    \textbf{3. + Weighted Sampling} & - & 0.50 $\pm$ 0.07 & 0.46 $\pm$ 0.10 & \underline{0.45} $\pm$ 0.17 & 0.80 $\pm$ 0.03 & 0.82 $\pm$ 0.03 & 0.87 $\pm$ 0.07 & 0.52 \\
    \textbf{4. + Binary Head} &  -  & 0.55 $\pm$ 0.09 & \textbf{0.52} $\pm$ 0.04 & \textbf{0.47} $\pm$ 0.15 & \underline{0.84} $\pm$ 0.03 & \underline{0.85} $\pm$ 0.05 & 0.87 $\pm$ 0.04 & \underline{0.53} \\
    \textbf{5. + Neutral Reference} & $k=1$ & \underline{0.58} $\pm$ 0.06 & 0.50 $\pm$ 0.04 & 0.43 $\pm$ 0.17 & \textbf{0.86} $\pm$ 0.03 & \textbf{0.86} $\pm$ 0.04 & \underline{0.88} $\pm$ 0.03 & \underline{0.53} \\
    \textbf{6. + Multi-shot Inference} & $k=3$ & \textbf{0.59} $\pm$ 0.06 & \underline{0.51} $\pm$ 0.04 & 0.44 $\pm$ 0.17 & \textbf{0.86} $\pm$ 0.02 & \textbf{0.86} $\pm$ 0.04 & \textbf{0.89} $\pm$ 0.03 & \textbf{0.54} \\
    \bottomrule
  \end{tabular}%
  }
\end{table*}
To efficiently adapt the frozen Dino-V3 backbone, we employ Low-Rank Adaptation (LoRA) with a rank of 8, freezing the base weights while injecting trainable adapters into the attention layers \cite{hu2022lora,simeoni2025dinov3}. This strategy is applied to both the pain and neutral reference encoders, thus reducing the parameter budget while preserving pre-trained representations.

\paragraph{Implementation Details.}
Input images are resized to $224 \times 224$. To preserve the generalization capabilities of the learned features, we freeze the backbone during both pretraining and 5-fold cross-validation. The optimization uses AdamW with learning rates: $1 \times 10^{-5}$ and weight decay of $1 \times 10^{-1}$. We employ a cosine annealing schedule to decay the learning rate to $1\%$ of its initial value over 100 epochs. The loss function weights are set to $\lambda_{PSPI} = 1.0$, $\lambda_{bin} = 1.0$, and $\lambda_{AU} = 0.1$. To prevent overfitting to the target domain, all architectural choices and hyperparameters were tuned exclusively on the held-out test set of the synthetic 3DPain dataset (following a 70/20/10 identity-disjoint split) rather than the UNBC-McMaster validation folds.

\paragraph{Evaluation Metrics.} 
Adhering to established benchmarks \cite{10970367,fiorentini2022fully}, we report the mean AUROC and F1-score across the five cross-validation folds on UNBC-McMaster. In contrast, the Pearson Correlation Coefficient (PCC) is computed on the aggregated validation folds to provide a holistic assessment of global ordinal consistency \cite{Rezaei2021}. We evaluate performance at clinically relevant PSPI thresholds ($\geq 2, \geq 3$) \cite{Rezaei2021}, while including $\geq 1$ for historical comparability.

\paragraph{Ablation Studies.}
To ensure generalization, all architectural decisions were finalized on the synthetic 3DPain dataset, with the limited real-world UNBC-McMaster dataset reserved strictly for evaluation via cross-validation.
As illustrated in Table \ref{tab:unbc_results}, baseline models trained from scratch exhibit limited generalization, due to the lack of diversity in the target domain. However, pre-training on synthetic data yields substantial performance gains across all architectures. We observe that transformer-based models benefit most significantly from this strategy; while ViTs possess higher representational capacity, they require the scale and variance provided by synthetic data to surpass the strong inductive biases of CNNs. This benefit is particularly pronounced at higher pain thresholds ($\ge 3$), where the uniform label distribution of 3DPain mitigates the scarcity of high-intensity samples in the skewed UNBC-McMaster dataset.

\paragraph{Component Analysis.}
Table \ref{tab:ablation_study} dissects the contribution of each architectural component, revealing a trade-off between decision-boundary sharpness and global ranking consistency.
The integration of a binary classification head maximizes F1-scores for high-intensity pain ($\ge 2, \ge 3$). By explicitly supervising the binary distinction between pain and non-pain, this auxiliary head calibrates the decision threshold, preventing the model from collapsing into the majority class (no-pain).
However, the inclusion of the neutral reference and multi-shot inference alters the performance profile. While this configuration causes a slight reduction in high-intensity F1 scores, attributable to a more conservative prediction bias, it significantly improves global discrimination metrics (AUROC $\ge 0.86$, PCC $\ge 0.54$). This indicates that referencing a neutral frame effectively disentangles subject identity from expression, fostering a smoother, monotonic latent space that favors ranking consistency over rigid binary classification.

\paragraph{Performance Comparison.} \begin{table}[t]
  \centering
  \caption{\textbf{State-of-the-Art Comparison:} Performance on UNBC-McMaster across Input Modality (\textbf{Input}), Validation Protocol (\textbf{Val.}), and Threshold (\textbf{Thr.}). Video methods benefit from temporal context. We evaluate \textbf{ViTPain} using 5-Fold CV.}
  \label{tab:sota_comparison}
  \resizebox{\linewidth}{!}{% Resize if too wide
  \begin{tabular}{@{}lllcccc@{}}
    \toprule
    \textbf{Method} & \textbf{Input} & \textbf{Val.} & \textbf{Thr.} & \textbf{AUROC $\uparrow$} & \textbf{F1 $\uparrow$} & \textbf{PCC $\uparrow$} \\
    \midrule
    % --- Video Based ---
    Rodriguez et al. \cite{Rodriguez2017} & Video & LOSO & $\ge 1$ & \textbf{0.93} & - & - \\
    Fiorentini et al. (ViViT-1) \cite{fiorentini2022fully} & Video & 5-Fold & $\ge 1$ & 0.86 & 0.55 & - \\
    Rau et al. (VSwin-T-1-TD) \cite{10970367} & Video & 5-Fold & $\ge 1$ & 0.87 & 0.59 & - \\
    \midrule
    % --- Image Based Baselines ---
    Lucey et al. (SVM) \cite{Lucey2011} & Image & LOSO & $\ge 1$ & 0.84 & - & - \\
    Kaltwang et al. (SVR with DCT) \cite{Kaltwang2012} & Image & LOSO & $\ge 1$ & - & 0.48 & 0.59 \\
    Parkhi et al. (Deep Face) \cite{BMVC2015_41} & Image & LOSO & $\ge 1$ & - & 0.59 & - \\
    Fiorentini et al. (ViT-1) \cite{fiorentini2022fully} & Image & 5-Fold & $\ge 1$ & 0.88 & 0.55 & - \\
    Rau et al. (Swin Transformer-0) \cite{10970367} & Image & 5-fold & $\ge 1$ & 0.80 & 0.53 & - \\
    Ertugrul et al. (AFAR) \cite{Ertugrul2019} & Image & 5-Fold & $\ge 2$ & - & 0.59 & - \\
    Rezaei et al. (PwCT, our impl.) \cite{Rezaei2021} & Image & 5-Fold & $\ge 2$ & 0.80 & 0.48 & 0.54 \\
    Rezaei et al. (PwCT w/ 3DPain) \cite{Rezaei2021} & Image & 5-Fold & $\ge 2$ & 0.82 & 0.48 & 0.55 \\
    \midrule
    % --- Your Method (Merged) ---
    \textbf{ViTPain} (\textit{k} = 3) (Ours) & Image & 5-Fold & $\ge 1$ & 0.86 $\pm$ 0.02 & \textbf{0.59 $\pm$ 0.06} & \multirow{3}{*}{0.54} \\
    \textbf{ViTPain} (\textit{k} = 3) (Ours) & Image & 5-Fold & $\ge 2$ & 0.86 $\pm$ 0.04 & 0.51 $\pm$ 0.04 & \\ 
    
    \textbf{ViTPain} (\textit{k} = 3) (Ours) & Image & 5-Fold & $\ge 3$ & \textbf{0.89 $\pm$ 0.03} & 0.44 $\pm$ 0.17 & \\ 
   
    \bottomrule
  \end{tabular}%
  }
\end{table} 
We evaluate our approach on UNBC-McMaster using 5-fold cross-validation with subject-independent splits. Our ViTPain approach achieves robust performance, significantly outperforming baselines. 
We contextualize these results against prior works, such as the CNN-LSTM \cite{Rodriguez2017}, which employ Leave-One-Subject-Out (LOSO) evaluation. While LOSO maximizes the training set size, we adopt a stratified 5-fold protocol to rigorously assess generalization variance across larger unseen subject groups. Furthermore, unlike video-based baselines that leverage temporal context, our approach operates purely on single frames. Even under this stricter evaluation setting, our method remains highly competitive with state-of-the-art video models. Our implementation of \cite{Rezaei2021} yields lower metrics than reported, as we exclude their private training data to ensure a fair comparison using public datasets.

\section{Conclusion}
\label{sec:conclusion}
We address data scarcity in pain assessment with 3DPain, a large-scale synthetic dataset of 82,500 expressions with precise AU annotations. To leverage this, we propose ViTPain, a reference-aware architecture contrasting neutral and expressive states to isolate pain features. Experiments on UNBC-McMaster confirm that synthetic pre-training significantly improves model generalization. While our approach bridges the data gap, challenges remain regarding the domain shift in skin texture realism and the modeling of long-term temporal dynamics. Future work will focus on enhancing photorealism and integrating temporal consistency. Ultimately, this framework provides a scalable foundation for facial expression analysis, offering a robust alternative to expensive clinical annotation.

% TODO
% ViTPain demonstrates the utility of 3DPain for augmentation, achieving 0.91 AUROC on UNBC-McMaster via cross-modal distillation and AU-specific query. Its dual-branch design supports pain classification and AU regression, offering a framework for facial expression analysis and reducing annotation bottlenecks in clinical research.

% References
% Make sure 'references.bib' is in the same folder
\bibliographystyle{splncs04}
\bibliography{references}

\end{document}